\documentclass[times,twocolumn,final]{elsarticle}

\usepackage{cag}
\usepackage{framed,multirow}
\usepackage{booktabs}

\usepackage{amssymb}
\usepackage{amsmath}
\journal{Computers \& Graphics}

\begin{document}

\begin{frontmatter}

%% Title, authors and addresses

%% use the tnoteref command within \title for footnotes;
%% use the tnotetext command for theassociated footnote;
%% use the fnref command within \author or \affiliation for footnotes;
%% use the fntext command for theassociated footnote;
%% use the corref command within \author for corresponding author footnotes;
%% use the cortext command for theassociated footnote;
%% use the ead command for the email address,
%% and the form \ead[url] for the home page:
%% \title{Title\tnoteref{label1}}
%% \tnotetext[label1]{}
%% \author{Name\corref{cor1}\fnref{label2}}
%% \ead{email address}
%% \ead[url]{home page}
%% \fntext[label2]{}
%% \cortext[cor1]{}
%% \affiliation{organization={},
%%             addressline={},
%%             city={},
%%             postcode={},
%%             state={},
%%             country={}}
%% \fntext[label3]{}

\title{Navigating Large-Pose Challenge for High-Fidelity Face Reenactment with Video Diffusion Model}

% \author{
% 	Mingtao Guo\textsuperscript{1}\hspace{3em} Guanyu Xing\textsuperscript{2}\hspace{3em} Yanli Liu\textsuperscript{1,3}$^{\dag}$ \\
% 	{\textsuperscript{1}National Key Laboratory of Fundamental Science on Synthetic Vision, } \\
% 	{Sichuan University, Chengdu, China} \\
%         {\textsuperscript{2}School of Cyber Science and Engineering, Sichuan University, Chengdu, China} \\
% 	{\textsuperscript{3}College of Computer Science, Sichuan University, Chengdu, China} \\
% 	{\tt\small mingtaoguo@stu.scu.edu.cn,\{xingguanyu, yanliliu\}@scu.edu.cn}
% }
\author[1]{Mingtao \snm{Guo}} 
\ead{mingtaoguo@stu.scu.edu.cn}

\author[2]{Guanyu \snm{Xing}} 
\ead{xingguanyu@scu.edu.cn}

\author[3]{Yanci \snm{Zhang}}
\ead{yczhang@scu.edu.cn}

\author[1,3]{Yanli \snm{Liu}\corref{cor1}}
\cortext[cor1]{Corresponding author: 
  Email: yanliliu@scu.edu.cn}
\emailauthor{yanliliu@scu.edu.cn}{Yanli Liu}

\address[1]{National Key Laboratory of Fundamental Science on Synthetic Vision, Sichuan University, Chengdu, 610065, China}
\address[2]{School of Cyber Science and Engineering, Sichuan University, Chengdu, 610065, China}
\address[3]{College of Computer Science, Sichuan University, Chengdu, 610065, China}

%% Abstract
\begin{abstract}
%% Text of abstract
Face reenactment aims to generate realistic talking head videos by transferring motion from a driving video to a static source image while preserving the source identity. Although existing methods based on either implicit or explicit keypoints have shown promise, they struggle with large pose variations due to warping artifacts or the limitations of coarse facial landmarks. In this paper, we present the Face Reenactment Video Diffusion model (FRVD), a novel framework for high-fidelity face reenactment under large pose changes. Our method first employs a motion extractor to extract implicit facial keypoints from the source and driving images to represent fine-grained motion and to perform motion alignment through a warping module. To address the degradation introduced by warping, we introduce a Warping Feature Mapper (WFM) that maps the warped source image into the motion-aware latent space of a pretrained image-to-video (I2V) model. This latent space encodes rich priors of facial dynamics learned from large-scale video data, enabling effective warping correction and enhancing temporal coherence. Extensive experiments show that FRVD achieves superior performance over existing methods in terms of pose accuracy, identity preservation, and visual quality, especially in challenging scenarios with extreme pose variations.

\end{abstract}

%%Graphical abstract
%\begin{graphicalabstract}
%\includegraphics{grabs}
%\end{graphicalabstract}

%%Research highlights
%\begin{highlights}
%\item Research highlight 1
%\item Research highlight 2
%\end{highlights}

%% Keywords
\begin{keyword}
%% keywords here, in the form: keyword \sep keyword

%% PACS codes here, in the form: \PACS code \sep code

%% MSC codes here, in the form: \MSC code \sep code
%% or \MSC[2008] code \sep code (2000 is the default)
Face Reenactment, Motion Extractor, Warping Feature Mapper, Motion-aware Latent Space, Video Diffusion Model.

\end{keyword}

\end{frontmatter}

%% Add \usepackage{lineno} before \begin{document} and uncomment 
%% following line to enable line numbers
% \linenumbers

%% main text
%%
\section{Introduction}
Face reenactment is the process of synthesizing a lifelike talking head video using a provided source image as a reference, guided by a driving video. In this synthesis, the resulting face maintains the identity attributes of the source image while adopting the pose and expressions from the driving video. Face reenactment has many valuable applications, including character role-playing, digital avatars, online education, video conferencing, etc.

Existing face reenactment methods have demonstrated remarkable capabilities in generating talking faces. Keypoints are typically employed by these methods to represent facial motion. According to whether the keypoints are learned automatically (implicit) or predefined (explicit), these methods can be categorized into two types: implicit keypoint-based methods and explicit keypoint-based methods. Among them, implicit keypoint-based methods \cite{siarohin2019first,hong2022depth,zhao2022thin,wang2021facevid2vid,guo2024liveportrait} learn keypoints from the source and driving images to estimate dense motion fields (e.g., optical flow). These fields are then used to warp the source image toward the pose and expression of the driving image. Finally, a generator \cite{goodfellow2014generative, karras2020analyzing, karras2021alias} inpaints occluded or degraded regions to produce the final frame. By providing dense and flexible motion guidance, these methods effectively capture fine-grained facial deformations. However, when there is a large pose discrepancy between the source and driving images, the limited identity and appearance information in a single source image often fails to support effective inpainting in severely warped regions, leading to degraded synthesis quality. In contrast, explicit keypoint-based methods \cite{ma2024follow, xie2024x, chen2024echomimic} rely on facial landmarks extracted from the driving video to generate pose maps as spatial conditions for the pre-trained Stable Diffusion model \cite{rombach2022high}. To maintain identity and appearance consistency with the source image, they further incorporate fine-grained texture and appearance features from the source using ReferenceNet \cite{hu2024animate}. Although these approaches can produce high-quality facial textures, they are fundamentally limited by the coarse nature of explicit facial landmarks \cite{lugaresi2019mediapipe, bulat2017far}, which primarily capture rigid facial contours. When handling large head poses (e.g., profile views), these rigid contours often result in overlapping or collapsed keypoints, causing the generated pose maps to lose meaningful facial structure and identity cues. Therefore, handling large pose variations remains a significant challenge for existing face reenactment methods.

To achieve high-fidelity face reenactment under large pose variations, we leverage implicit facial keypoints to represent facial motion and use them to warp the source image toward the target pose and expression. To address warping-induced degradation, we observe that image-to-video (I2V) models trained on large-scale video datasets are highly effective at synthesizing realistic and temporally coherent facial dynamics—such as head movements, speech, and blinking—while reliably preserving identity and appearance consistency, even under extreme pose variations. Therefore, our key insight is to exploit the I2V model’s motion-aware latent feature space to reconstruct regions degraded by warping, enabling temporally coherent video generation that faithfully preserves source identity while recovering fine-grained details lost during the warping process. 

In this paper, we propose a \textbf{F}ace \textbf{R}eenactment \textbf{V}ideo \textbf{D}iffusion model (\textbf{FRVD}) for high-fidelity face reenactment under large pose variations. Our model first employs a Motion Extractor to extract implicit keypoints from both the source and driving images, which serve as fine-grained representations of facial motion. These keypoints are then used in a warping module to align the motion of the source image with that of the driving image. To recover regions degraded during the warping process, we introduce a Warping Feature Mapper (WFM) that maps features from the warped source image into the motion-aware latent space of a pretrained I2V model. This latent space, learned from large-scale video data, encodes rich spatiotemporal priors of facial dynamics, enabling the model to perform effective warping correction. By leveraging these priors, the WFM facilitates high-quality reconstruction of facial details while preserving both identity and temporal coherence.

Our main contributions are summarized as follows:
\begin{itemize}
	\item We propose a \textbf{F}ace \textbf{R}eenactment \textbf{V}ideo \textbf{D}iffusion model (\textbf{FRVD}) to address the challenge of face reenactment under large pose variations, overcoming the limitation of existing methods, which typically produce satisfactory results only when the pose of the source image closely matches that of the driving image.
	\item We introduce the Warping Feature Mapper (WFM), which maps the warped source image into the motion-aware latent space of a pretrained image-to-video (I2V) model. This allows the model to leverage its prior knowledge to reconstruct degraded regions, thereby enabling high-fidelity face reenactment under large pose variations.
	\item Extensive experiments demonstrate that FRVD outperforms state-of-the-art methods, achieving significant improvements in pose accuracy, identity preservation, and overall video quality.
\end{itemize}

\section{Related Work}
\textbf{Implicit-Keypoints-Based Face Reenactment Methods.} Contemporary face reenactment approaches leveraging implicit keypoint representation \cite{siarohin2019first,hong2022depth,zhao2022thin,wang2021facevid2vid,guo2024liveportrait} eliminate the need for prior knowledge of driving subjects during model training. Notably, the FOMM \cite{siarohin2019first} establishes a theoretical framework through first-order Taylor expansions around latent keypoints, implementing local affine transformations to approximate complex facial motions. This foundational work demonstrates significant performance improvements in motion transfer fidelity. Building upon this foundation, DaGAN \cite{hong2022depth} introduces a self-supervised paradigm for fine-grained pixel-level depth estimation, enabling enhanced 3D facial structure perception and high-frequency spatial detail preservation. In contrast, LivePortrait \cite{guo2024liveportrait} proposes a novel motion cue disentanglement mechanism that implicitly captures and transfers holistic facial dynamics—including head pose and expression variations—from driving videos while maintaining content consistency. Despite these advancements, fundamental limitations remain: existing methods still suffer from performance degradation under extreme pose variations, often resulting in geometric distortions and unrealistic texture artifacts caused by the warping module.

\textbf{Explicit-Keypoints-Based Face Reenactment Methods.} In contrast to implicit-keypoints-based methods, explicit keypoint-driven approaches \cite{ma2024follow, xie2024x, chen2024echomimic, guo2025high} typically rely on existing facial landmark detection models \cite{lugaresi2019mediapipe, bulat2017far} to extract keypoints from each frame of the driving video. These keypoints are then used to construct facial contour maps, which serve as pose guidance for the generation model. The pose guider directs the synthesis of facial images with corresponding head poses. Meanwhile, identity and appearance information from the source image is preserved by leveraging spatial attention mechanisms between features extracted by the ReferenceNet \cite{hu2024animate}  and those in the UNet of the Stable Diffusion model \cite{rombach2022high}. Additionally, diffusion-based face reenactment methods \cite{kligvasser2024anchored, bounareli2024diffusionact} also leverage facial contours derived from explicit keypoints to guide the diffusion model in generating faces with the desired poses. However, under extreme poses, explicit-keypoints-based approach often fails to preserve facial structure in the contour map, resulting in ineffective guidance and noticeable distortions in the reenacted face.

\textbf{Image-to-Video Diffusion Models. }Image-to-video diffusion models \cite{blattmann2023stable, yang2024cogvideox, wan2025} aim to generate a video from a single reference image, where the first frame is identical to the input image, and subsequent frames maintain consistent foreground and background while the motion is guided by user-provided textual descriptions. To ensure temporal consistency across frames, the Stable Video Diffusion (SVD) \cite{blattmann2023stable} model extends Stable Diffusion \cite{rombach2022high} by introducing a temporal attention module. To address the limitation of separate spatial and temporal attention—particularly the failure to track fast-moving objects—CogVideoX \cite{yang2024cogvideox} integrates spatial attention, cross-attention, and temporal attention \cite{guoanimatediff} into a unified self-attention mechanism, enabling stronger semantic understanding and temporal coherence. However, this unified attention design significantly increases the number of tokens, leading to high computational costs and low inference efficiency. Considering these trade-offs, we adopt SVD as the backbone for our image-to-video generation model.

\section{Methodology}
\begin{figure*}
	\centering
	\includegraphics[width=0.91\textwidth]{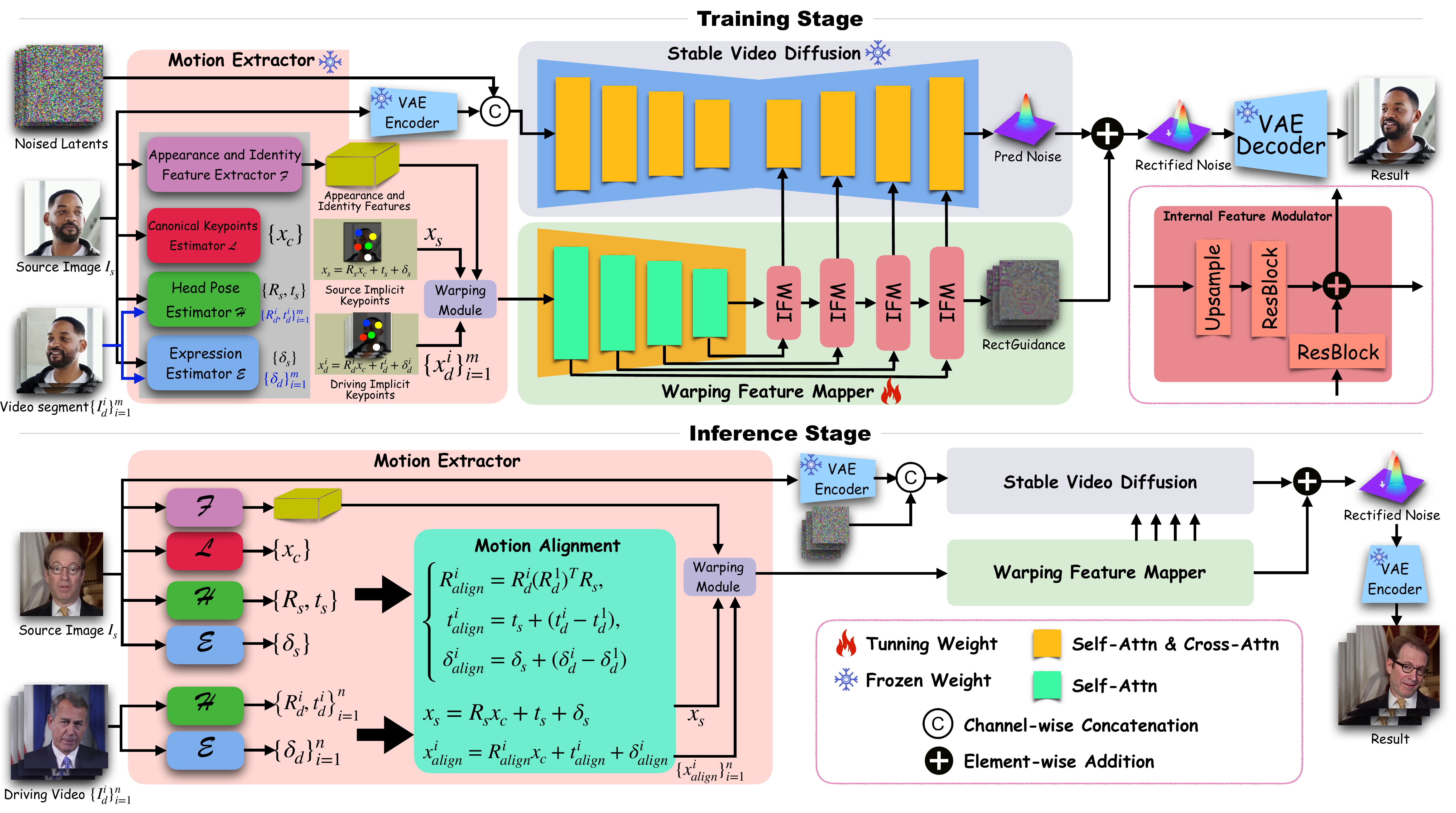}
	\caption{Our face reenactment framework comprises two stages: (1) Training stage: We begin by employing the Motion Extractor to extract pose and expression coefficients from both the source image and the driving video, while simultaneously encoding appearance and identity features from the source image. These motion coefficients are then used to warp the source image features via the Warping Module, aligning them with the motion of the driving video. The warped features are further encoded by the Warping Feature Mapper and modulated by the Internal Feature Modulator before being passed to the Stable Video Diffusion (SVD) model. Leveraging the identity and appearance priors of the source image inherent in SVD, the model performs warping correction in the motion-aware latent space to reconstruct regions distorted or missing due to warping. (2) Inference stage: To support cross-identity face reenactment, a Motion Alignment Module is introduced into the pipeline, enabling the model to generalize to unseen identities.
}
	\label{fig2}
\end{figure*}
\subsection{Overview}
Our method takes a source image and a driving video as input, and reenacts the source face to match the pose in each frame of the driving video. As illustrated in Fig. \ref{fig2}, our face reenactment framework consists of two stages. During the training stage, we first employ a Motion Extractor to extract pose and expression coefficients from both the source image and the driving video. The source image is then warped to match the pose and expression of the driving video, ensuring spatial alignment with each frame (Sec. \ref{mew}). Next, the warped source image is encoded by a Warping Feature Mapper (WFM). At each layer of the WFM, the extracted features are modulated by an Internal Feature Modulator (IFM). These modulated features are then fed into the Stable Video Diffusion (SVD) model \cite{blattmann2023stable}. By leveraging the identity and appearance priors of the source image embedded in SVD, we perform warping correction within its motion-aware latent space to reconstruct the missing regions of the warped source image (Sec. \ref{iimwr}). During the inference stage, we introduce an additional Motion Alignment Module into the Motion Extractor to support cross-identity face reenactment (Sec. \ref{cfr}).
\subsection{Motion Extraction and Warping} 
\label{mew}
The core objective of face reenactment is to transfer the pose and expression from a driving video to a source image, thereby transforming a static source image into a dynamic face that maintains the identity of the source image while aligning with the pose and expression of the driving video. To achieve this, the primary task is to disentangle the pose, expression, identity, and appearance from the source image.

As illustrated in Fig. \ref{fig2}, the training stage of the motion extractor, we first employ the appearance and identity feature extractor $\mathcal{F}$ to extract features from the source image, denoted as $\mathbf{F}_s$. Subsequently, the canonical keypoint estimator $\mathcal{L}$ is used to extract facial keypoints in the canonical space from the source image. Next, the head pose estimator $\mathcal{H}$ estimates the pose of the source image’s face, represented by a rotation matrix $\mathbf{R}_s$ and a translation matrix $\mathbf{t}_s$. For each frame of the driving video, the facial pose is also estimated and denoted as $\left\{\mathbf{R}_d^i\right\}_{i = 1}^m$ and $\left\{\mathbf{t}_d^i\right\}_{i = 1}^m$, where $m$ represents the total number of frames in the driving video. Finally, the expression estimator $\mathcal{E}$ is utilized to estimate the facial expression coefficients for both the driving video and the source image, denoted as $\left\{\boldsymbol{\delta}_d^i\right\}_{i = 1}^m$ and $\boldsymbol{\delta}_s$, respectively.

Based on these estimated parameters, the facial keypoints for the source image and each frame of the driving video are computed using Eq. (\ref{eq:1}),
\begin{equation}
    \begin{cases}
        \mathbf{x}_s = \mathbf{R}_s \mathbf{x}_c + \mathbf{t}_s + \boldsymbol{\delta}_s \\
        \mathbf{x}_d^i = \mathbf{R}_d^i \mathbf{x}_c + \mathbf{t}_d^i + \boldsymbol{\delta}_d^i
    \end{cases}
    \label{eq:1}
\end{equation}
finally, following OSFV \cite{wang2021facevid2vid}, we utilize the source image keypoints $\mathbf{x}_s$ and the driving video frame keypoints $\mathbf{x}_d^i$ in the warping module to warp the source image features $\mathbf{F}_s$. This process is formulated as $f_w: (\mathbf{x}_s,\mathbf{x}_d^i,\mathbf{F}_s) \rightarrow \mathbf{F}_s^w$, where $\mathbf{F}_s^w$ represents the warped version of $\mathbf{F}_s$, ensuring that its pose is aligned with the driving frame.
\subsection{Warping Correction in Motion-aware Latent Space} 
\label{iimwr}
Using the Motion Extractor, we warp the source image in the feature space to align its pose and expression with those of the driving image. However, the warping process inevitably introduces information loss in the affected regions of the feature map. This problem becomes particularly pronounced when there is a significant pose discrepancy between the source and driving images, leading to large-scale facial distortions and severe identity degradation.

To address the distortion and identity loss introduced by warping, our key idea is to extract identity and appearance features from the source image and utilize them to restore the regions degraded during the warping process. We observe that image-to-video (I2V) models are inherently capable of synthesizing temporally continuous frames from a single input image, enabling realistic image-to-video generation. Given a facial image, I2V models can predict natural motions such as head turns, speech, and blinking, while maintaining consistent identity and appearance across frames. This makes them particularly well-suited for handling faces under varying poses. We exploit this property by leveraging the perceptual capabilities of a pre-trained I2V model to restore warped regions at the feature level, ensuring that the reconstructed face remains consistent with the source identity while generating temporally coherent driving videos.

Specifically, we propose a Warping Feature Mapper (WFM), as illustrated in Fig. \ref{fig2}. We first feed the warped feature $\mathbf{F}_s^w \in \mathbb{R}^{C\times H\times W}$ into WFM, where it is encoded by the WFM Encoder (WFMEnc), denoted as $[\mathbf{F}_s^{(1)}, …, \mathbf{F}_s^{(i)}, …] = \text{WFMEnc}(\mathbf{F}_s^w)$, where each $\mathbf{F}_s^{(i)}\in \mathbb{R}^{C^{(i)}_s\times H^{(i)}_s\times W^{(i)}_s}$ denotes the feature representation at the $i$-th scale. The encoded features $\mathbf{F}_s^{(i)}$ from each encoder layer are then passed to the corresponding Internal Feature Modulator (IFM) for feature modulation: $\mathbf{F}_{s\_m}^{(i)} = \text{IFM}_i(\mathbf{F}_s^{(i)})$, where $\mathbf{F}_{s\_m}^{(i)}\in \mathbb{R}^{C^{(i)}\times H^{(i)}\times W^{(i)}}$ represents the modulated features at the $i$-th scale. The modulated features $\mathbf{F}_{s\_m}^{(i)}$ are subsequently fed into a pre-trained I2V model. Leveraging the I2V model’s perceptual ability to encode identity and appearance from the source image, we use it to restore the features degraded by warping. Let $\mathbf{F}^{(j)}$ denote the feature map from the $j$-th layer of the I2V model, which contains rich identity and appearance cues from the source image. The tensor $\mathbf{F}_{s\_m}^{(i)}$ shares the same shape as $\mathbf{F}^{(j)}$. We fuse $\mathbf{F}^{(j)}$ with $\mathbf{F}_{s\_m}^{(i)}$ via element-wise addition to inject identity-aware information into the warped features: $\mathbf{F}_{fuse}^{(j)} = \mathbf{F}^{(j)} + \mathbf{F}_{s\_m}^{(i)}$. The fused feature $\mathbf{F}_{fuse}^{(j)}$ is then fed into the next layer of the I2V model, and this process continues recursively. In our framework, the I2V model is implemented using the SVD model \cite{blattmann2023stable}.

Additionally, to enhance SVD’s ability to capture the global appearance characteristics of the source image, we introduce a rectified guidance signal predicted by the final IFM layer. This signal shifts the mean of the Gaussian distribution output by SVD, enabling more accurate modeling of the source image’s appearance. The training objective is defined in Eq. (\ref{eq:2}).
\begin{equation}
\mathbf{Loss}=\mathbb{E}\left[\|\boldsymbol{\epsilon}-\boldsymbol{\epsilon}_{\boldsymbol{\theta}}\left( \sqrt{\bar{\alpha}_t}\mathbf{z}_0-\sqrt{1-\bar{\alpha}_t}\boldsymbol{\epsilon},\mathbf{F}_c,t\right)-\mathbf{r}\|\right]
    \label{eq:2}
\end{equation}
Here, $\boldsymbol{\epsilon}$ represents noise sampled from a standard normal distribution, while $\mathbf{z}_0$ denotes the latent representation of the driving image (i.e., the target image) obtained from the VAE \cite{esser2021taming}. $\boldsymbol{\epsilon}_\theta$ refers to the backbone of the SVD model. The operation $\left[\mathbf{F}_c,\mathbf{r}\right]=WFM\left(\mathbf{F}_s^w\right)$ denotes the output of the WFM, where $\mathbf{F}_c=[\mathbf{F}_{s\_m}^{(1)},...,\mathbf{F}_{s\_m}^{(i)},...]$ is the modulated feature output from WFM, and $\mathbf{r}$ is the rectified guidance used to shift the mean of the Gaussian distribution predicted by SVD. The term ${\bar{\alpha}}_t=\prod_{i=1}^{t}\left(1-\beta_i\right)$, where $\beta_i$ represents the noise strength coefficient.

\subsection{Cross-identity Face Reenactment}
\label{cfr}
Face reenactment aims to drive an image with a video of another identity. During training, a frame is randomly selected as the source image, and a video segment is cropped as the driving video. The model processes the source image to generate a video matching the driving video, enabling self-supervised learning.

However, the goal of face reenactment is to drive a face image of one identity using a face video of another identity. To prevent identity leakage from the driving video, we employ the Motion Alignment Module to align the pose and expression of the driving frames with those of the source image. Our approach computes the facial motion for each frame of the driving video, using the first frame as a reference. We then calculate the motion differences between each frame and the reference, obtaining a relative motion sequence. This sequence is applied to the source image, producing a face reenactment video where the initial motion matches the source image, and subsequent pose and expression changes follow the relative sequence smoothly and consistently. Specifically, we first use the Motion Extractor to compute the rotation matrices $\left\{\mathbf{R}_d^i\right\}_{i=i}^m$, translation vectors $\left\{\mathbf{t}_d^i\right\}_{i=i}^m$, and expression coefficients $\left\{\boldsymbol{\delta}_d^i\right\}_{i=i}^m$ for each frame of the driving video. Similarly, we extract the source image’s rotation matrix $\mathbf{R}_s$, translation vector $\mathbf{t}_s$, and expression coefficients $\boldsymbol{\delta}_s$.

Using the first frame’s rotation matrix as a reference, we compute the aligned rotation matrix, translation vector and expression coefficients as Eq. (\ref{eq:3}).
\begin{equation}
    \begin{cases}
\mathbf{R}_{align}^i=\mathbf{R}_d^i\left(\mathbf{R}_d^1\right)^T\mathbf{R}_s \\
\mathbf{t}_{align}^i=\mathbf{t}_s+\left(\mathbf{t}_d^i-\mathbf{t}_d^1\right)  \\
\boldsymbol{\delta}_{align}^i=\boldsymbol{\delta}_s^i+\left(\boldsymbol{\delta}_d^i-\boldsymbol{\delta}_d^1\right)
    \end{cases}
    \label{eq:3}
\end{equation}
Using these aligned pose and expression parameters, we compute the facial keypoints for both the source image and the driving video, as described in Eq (\ref{eq:4}).
\begin{equation}
    \begin{cases}
\mathbf{x}_s=\mathbf{R}_s\mathbf{x}_c+\mathbf{t}_s+\boldsymbol{\delta}_s \\
\mathbf{x}_{align}^i=\mathbf{R}_{align}^i\mathbf{x}_c+\mathbf{t}_{align}^i+\boldsymbol{\delta}_{align}^i
    \end{cases}
    \label{eq:4}
\end{equation}
We then employ these facial keypoints to warp the source image features, denoted as $\mathbf{F}_s^w = f_w(\mathbf{x}_s, \mathbf{x}_{align}, \mathbf{F}_s)$. The warped features are subsequently encoded by the WFM, $\left[\mathbf{F}_c,\mathbf{r}\right]=WFM\left(\mathbf{F}_s^w\right)$, and further processed by the SVD model. SVD leverages the DDIM reverse diffusion process \cite{song2020denoising} to perform warping correction, effectively restoring regions lost during the warping operation. Additionally, we adopt a classifier-free guidance mechanism to control the strength of the correction, as formulated in Eq. (\ref{eq:5}), where adjusting the value of $w$ modulates the degree of restoration in the warped regions.
\begin{equation}
    \boldsymbol{\epsilon}_\theta(\mathbf{z}_t)=w\cdot(\boldsymbol{\epsilon}_\theta(\mathbf{z}_t,\mathbf{F}_c)-\boldsymbol{\epsilon}_\theta(\mathbf{z}_t,\phi))+\boldsymbol{\epsilon}_\theta(\mathbf{z}_t,\mathbf{F}_c)+\mathbf{r}
    \label{eq:5}
\end{equation}

\section{Experiments}
\subsection{Implementation Details}
\noindent\textbf{Datasets. }We train our model using the VFHQ \cite{xie2022vfhq} datasets. For fair evaluation, we conduct self-reenactment and cross-identity reenactment experiments on the HDTF \cite{zhang2021flow} and CelebV-HQ \cite{zhu2022celebv} dataset, analyzing results both quantitatively and qualitatively.

\noindent\textbf{Training Details. }During training, we sample a 14-frame video sequence to ensure temporal consistency within SVD’s temporal attention layer, with each frame at a resolution of $512 \times 512$. The weights of the Motion Extractor and SVD are kept fixed, while only the Warping Feature Mapper and Internal Feature Modulator are updated. The model is trained for 30,000 iterations with a batch size of 8, using gradient accumulation and gradient checkpointing to manage memory consumption. Optimization is performed using 8-bit Adam \cite{kingma2014adam} with a learning rate of $1 \times 10^{-5}$ on a single A6000 GPU.

\noindent\textbf{Inference Details. }During inference, we input 14-frame sequences into the model with a 6-frame overlap, following \cite{zhang2024mimicmotion} to ensure temporal consistency. We use DDIM sampling with 30 steps and a guidance scale of 2.5. On an RTX 4090, generating a 100-frame video takes about 4 minutes.
\subsection{Metrics and Comparisons}
\noindent\textbf{Evaluation Metrics. }To evaluate our method, we conduct self-reenactment and cross-identity reenactment experiments on the HDTF \cite{zhang2021flow} dataset. For self-reenactment, we assess the similarity between the reenacted results and the driving video using Mean Absolute Error (L1), Peak Signal-to-Noise Ratio (PSNR), and Structural Similarity (SSIM) \cite{wang2004image}. Perceptual error is measured with LPIPS \cite{zhang2018unreasonable}, which uses a pre-trained AlexNet \cite{krizhevsky2012imagenet} model. Additionally, identity preservation (ID) is assessed using an ArcFace-based \cite{deng2019arcface} face recognition model. For cross-identity reenactment, we use ID to compare the reenacted results with the source image. Average Pose Distance (POSE) is computed by detecting facial keypoints in both the reenacted results and the driving video and calculating the keypoint error. To evaluate expression score (EXP), we follow \cite{hong2022depth} to measure expression similarity between the reenacted results and the driving video. Additionally, we utilize a no-reference video quality assessment model \cite{wu2022fasterquality} to evaluate the video quality (VQ) of the reenacted results, and adopt Fréchet Inception Distance (FID) \cite{heusel2017gans} and Fréchet Video Distance (FVD) \cite{unterthiner2018towards} to measure visual fidelity.Beyond these objective metrics, we also conduct a user study to further evaluate the quality of the face reenactment results. The study assesses four key aspects: pose accuracy (POSE-User), expression realism (EXP-User), identity preservation (ID-User), and video quality (VQ-User). Each dimension is rated on a five-point Likert scale: 1 (Poor), 2 (Fair), 3 (Average), 4 (Good), and 5 (Excellent). A total of 11 participants took part in the user study.

\noindent\textbf{Comparative Methods. }We conduct a comparative analysis between our method and seven state-of-the-art face reenactment methods: OSFV \cite{wang2021facevid2vid}, TPSMM \cite{zhao2022thin}, LivePortrait \cite{guo2024liveportrait}, FADM \cite{zeng2023face}, AniPortrait \cite{wei2024aniportrait}, Echomimic \cite{chen2024echomimic}, and FollowYourEmoji \cite{ma2024follow}. All methods are evaluated on the HDTF and CelebV-HQ datasets.
\subsection{Quantitative Evaluation}

\begin{table*}[t!]
    \centering
    \caption{Quantitative comparison of self-reenactment and cross-identity reenactment on the HDTF video dataset. We evaluate seven state-of-the-art face reenactment methods: OSFV, TPSMM, LivePortrait, FADM, AniPortrait, Echomimic, and FollowYourEmoji. The best scores are highlighted in bold, and the second-best are underlined.}
    % \vspace{-2mm}
    \label{tab:compare}
    \scalebox{0.98}
    {\begin{tabular}{cccccc||cccccc}
        \hline
        &\multicolumn{5}{c}{Self-Reenactment}&\multicolumn{6}{c}{Cross-identity Reenactment} \\
        \cmidrule(r){2-6} \cmidrule(r){7-12}
        Methods & L1$\downarrow$ & PSNR$\uparrow$ & SSIM$\uparrow$ & LPIPS$\downarrow$ & ID$\uparrow$ & POSE$\downarrow$ & EXP$\uparrow$ & ID$\uparrow$ & FID$\downarrow$ & FVD$\downarrow$ & VQ$\uparrow$ \\
        \hline
        % FOMM \cite{siarohin2019first} & 0.0334 & 24.88 & 0.8077 & 0.1953  & 0.8487 & 3.615 & 0.6405 & 0.8921 & 27.64 & 211.2 & 0.5905 \\
        % DaGAN \cite{hong2022depth} & 0.0279 & 27.023 & 0.8523 & 0.1623  & 0.8556 & \textbf{2.088} & 0.6506 & 0.8962 & 40.74 & 172.9 & 0.6224 \\
        OSFV \cite{wang2021facevid2vid} & 0.0303 & 26.221 & 0.8476 & 0.1531 & 0.8584 & 3.239 & 0.6456 & \underline{0.9127} & 20.85 & 169.4 & 0.6318 \\
        TPSMM \cite{zhao2022thin} & \underline{0.0256} & \underline{27.582} & \underline{0.8692} & 0.1550 &  0.8662 & \underline{2.806} & 0.6495 & 0.8962 & 22.39 & 175.2 & 0.6173 \\
        LivePortrait \cite{guo2024liveportrait} & 0.0418 & 22.788 & 0.7746 & 0.1222 & \textbf{0.8967} & 6.159 & 0.6384 & \textbf{0.9294} & 25.07 & \underline{142.8} & 0.7739 \\
        FADM \cite{zeng2023face} & 0.0359 & 25.330 & 0.8348 & 0.1765 & 0.8463 & 18.25 & 0.6256 & 0.8994 & 22.29 &164.6 &0.6275 \\
        AniPortrait \cite{wei2024aniportrait} & 0.0412 & 21.518 & 0.7702 & 0.1680 & 0.8506 & 4.375 & 0.6558 & 0.8854 & 18.35 &277.5 &\underline{0.7954} \\
        Echomimic \cite{chen2024echomimic} & 0.0353 & 24.166 & 0.8096 & \underline{0.1020} & \underline{0.8674} & 19.29 & \underline{0.6685} & 0.7635 & 19.88 &399.4 &0.7192 \\
        FollowYourEm. \cite{ma2024follow} & 0.0358 & 23.753 & 0.7951 & 0.1077 & 0.8535 & 4.341 & 0.6625 & 0.8976 & \underline{15.75} &164.1 &0.7577 \\

        \hline
        Ours & \textbf{0.0226} & \textbf{27.708} & \textbf{0.8702} & \textbf{0.0760} & 0.8570 & \textbf{2.695} & \textbf{0.6710} & 0.8975 &\textbf{14.73} &\textbf{140.8} &\textbf{0.8061} \\
        \hline
    \end{tabular}}
    % \vspace{-4mm}
\end{table*}

\begin{table}[t!]
    \centering
    \caption{User study comparison of cross-identity reenactment on the HDTF video dataset. We conduct a user study to evaluate the perceptual quality of seven state-of-the-art face reenactment methods. Participants were asked to score the results of each method based on pose accuracy, expression realism, identity preservation, and overall video quality. The highest-rated scores are highlighted in bold, and the second-highest are underlined.}
    % \vspace{-2mm}
    \label{tab:compare_user}
    \scalebox{0.75}
    {\begin{tabular}{ccccc}
        \hline
        Methods & POSE-User$\uparrow$ & EXP-User$\uparrow$ & ID-User$\uparrow$ & VQ-User$\uparrow$ \\
        \hline
        OSFV \cite{wang2021facevid2vid} & 3.246 & \underline{3.560} & 3.812 & 3.650 \\
        TPSMM \cite{zhao2022thin} & 2.833 & 2.626 & 3.020 & 3.293 \\
        LivePortrait \cite{guo2024liveportrait} & \underline{4.211} & 3.478 & \underline{4.118} & \underline{3.800} \\
        FADM \cite{zeng2023face} & 2.833 & 2.502 & 2.823 & 2.414 \\
        AniPortrait \cite{wei2024aniportrait} & 3.522 & 2.375 & 3.011 & 2.325  \\
        Echomimic \cite{chen2024echomimic} & 3.190 & 2.612 & 2.820 & 2.375 \\
        FollowYourEm. \cite{ma2024follow} & 3.012 & 2.713 & 2.491 & 2.582 \\
        \hline
        Ours & \textbf{4.372} & \textbf{4.023} & \textbf{4.375} & \textbf{3.835} \\
        \hline
    \end{tabular}}
    % \vspace{-4mm}
\end{table}

\begin{figure*}
	\centering
	\includegraphics[width=0.99\textwidth]{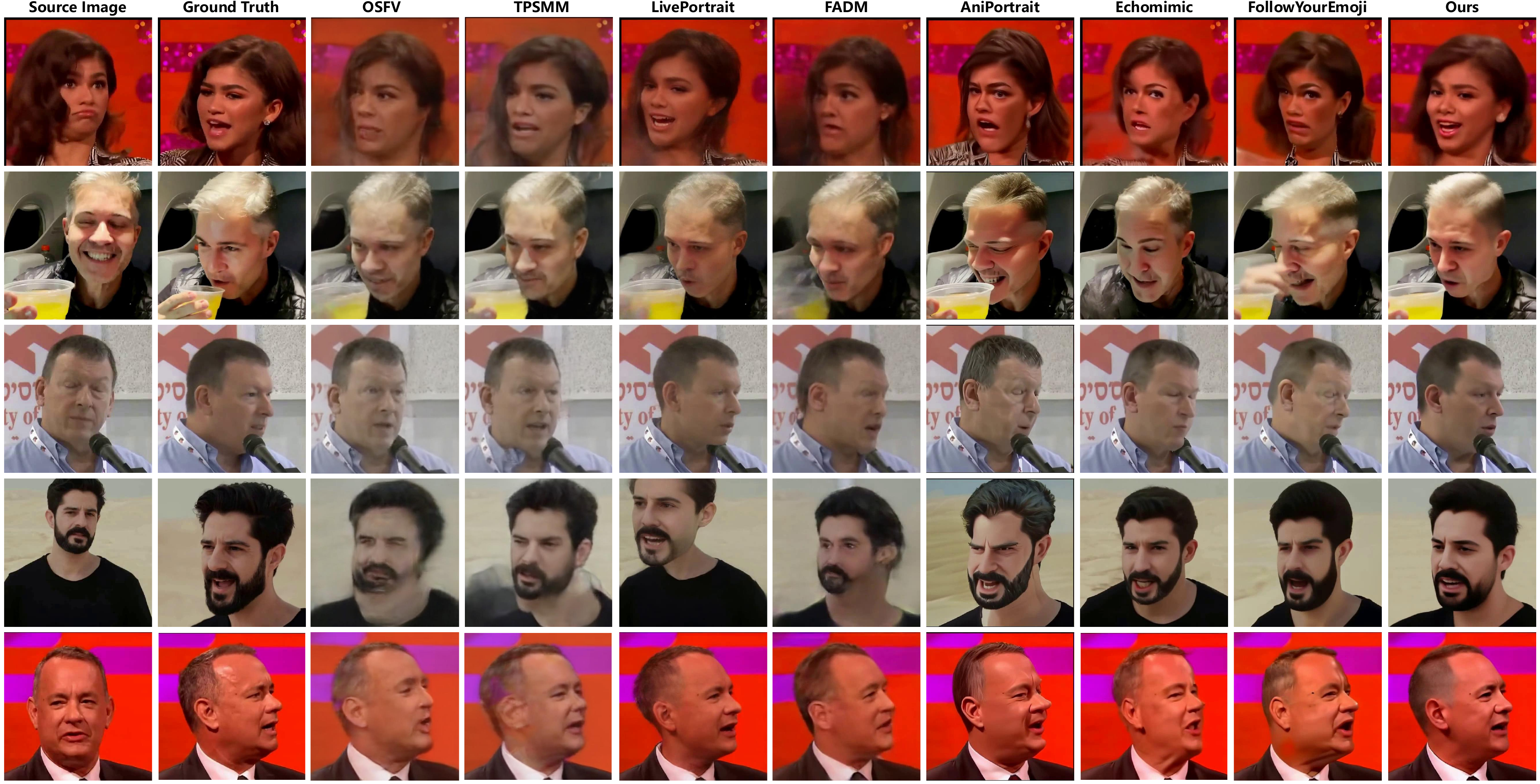}
	\caption{Self-reenactment qualitative comparison with state-of-the-art methods including OSFV \cite{wang2021facevid2vid}, TPSMM \cite{zhao2022thin}, LivePortrait \cite{guo2024liveportrait}, FADM \cite{zeng2023face}, AniPortrait \cite{wei2024aniportrait}, Echomimic \cite{chen2024echomimic}, and FollowYourEmoji \cite{ma2024follow}. The first column shows the source image, the second column presents the driving image (ground truth), and the remaining columns display the reenacted results. Our method delivers more realistic outcomes, especially under challenging conditions such as extreme facial poses.}
	\label{fig_self}
\end{figure*}
We compare our method with seven state-of-the-art face reenactment approaches: OSFV \cite{wang2021facevid2vid}, TPSMM \cite{zhao2022thin}, LivePortrait \cite{guo2024liveportrait}, FADM \cite{zeng2023face}, AniPortrait \cite{wei2024aniportrait}, Echomimic \cite{chen2024echomimic}, and FollowYourEmoji \cite{ma2024follow}. To provide a comprehensive evaluation, we conduct experiments under both self-reenactment and cross-identity reenactment settings.

For self-reenactment, we use the first frame of each video as the source image and the subsequent frames as driving frames. The objective is to generate a reenacted frame that aligns with the corresponding driving frame, allowing us to use the driving frame as ground truth for evaluation. Additionally, to assess the robustness of our method, we conduct cross-identity reenactment, where a face video of one identity is used to drive a face image of another identity.

As shown in Table \ref{tab:compare}, in the self-reenactment, our method outperforms all other approaches across multiple metrics, including L1, PSNR, SSIM, and LPIPS. Notably, our method achieves an 11.7\% lower pixel-wise error (L1) and a 25.5\% lower perceptual error (LPIPS) compared to the second-best method. However, in terms of identity preservation (ID), our method is slightly inferior to LivePortrait. For the cross-identity reenactment, our method achieves the best performance in POSE, EXP, FID, and FVD metrics. Specifically, it reduces POSE error by 4.0\% and FID by 6.5\% compared to the second-best method. Additionally, LivePortrait outperforms our model in terms of the ID metric, primarily due to its use of a large-scale private facial video dataset exceeding 16 million frames and the explicit identity supervision provided by the ArcFace \cite{deng2019arcface} face recognition model, which significantly enhances identity preservation. However, except for the ID metric, our method achieves the best performance across all other metrics, demonstrating overall superiority over other state-of-the-art methods.

Since cross-identity reenactment lacks ground-truth data for direct evaluation, relying solely on no-reference image quality metrics may lead to an incomplete assessment. While objective metrics such as PSNR, FID, and LPIPS are commonly reported, the quality of facial generation often depends heavily on human perception—particularly with respect to expression realism, identity consistency, and overall naturalness. A user study is therefore essential, especially in cases where metric differences are small but perceptual differences are significant.

To this end, we conducted a comprehensive user study to evaluate the effectiveness of our method. As shown in Table \ref{tab:compare_user}, our approach achieves the highest ratings across all four perceptual dimensions: pose (POSE-User), expression (EXP-User), identity preservation (ID-User), and video quality (VQ-User). Notably, our method outperforms the second-best method by 13\% in expression realism and by 6.2\% in identity preservation. These results indicate that, although our method may not lead in all objective identity metrics, it consistently produces superior perceptual quality, demonstrating its effectiveness in generating visually compelling reenactment results.
\begin{figure*}
	\centering
	\includegraphics[width=0.99\textwidth]{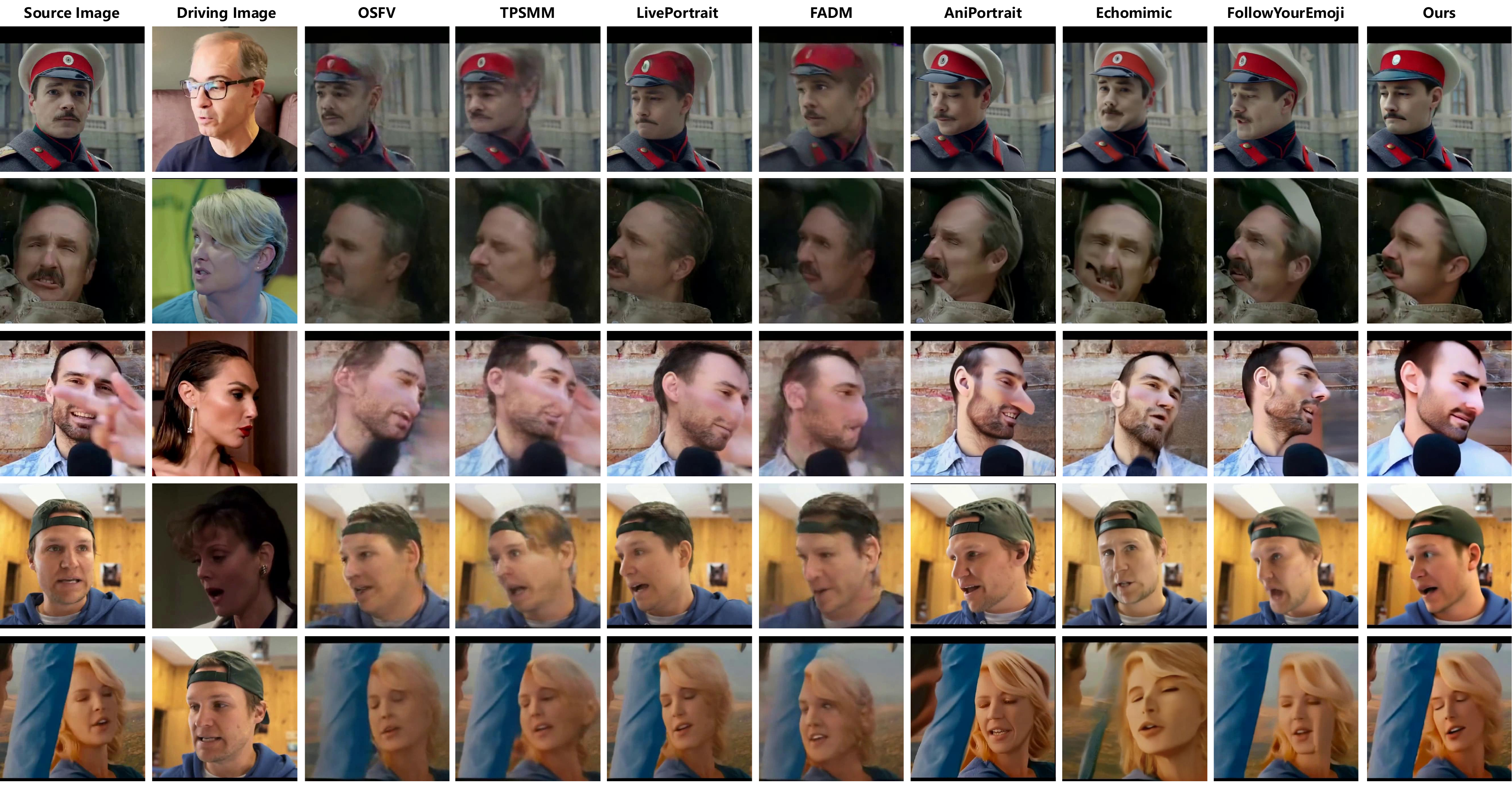}
	\caption{Cross-identity reenactment qualitative comparison with state-of-the-art methods, including OSFV \cite{wang2021facevid2vid}, TPSMM \cite{zhao2022thin}, LivePortrait \cite{guo2024liveportrait}, FADM \cite{zeng2023face}, AniPortrait \cite{wei2024aniportrait}, Echomimic \cite{chen2024echomimic}, and FollowYourEmoji \cite{ma2024follow}. The first column shows the source image, the second presents the driving image, and the remaining columns display reenacted results. Our method produces more realistic outputs}
	\label{fig_cross}
\end{figure*}
\subsection{Qualitative Evaluation}
We present a qualitative comparison between our method and recent state-of-the-art face reenactment approaches on the CelebV-HQ \cite{zhu2022celebv} dataset, covering both self-reenactment and cross-identity reenactment settings. The comparison includes OSFV \cite{wang2021facevid2vid}, TPSMM \cite{zhao2022thin}, LivePortrait \cite{guo2024liveportrait}, FADM \cite{zeng2023face}, AniPortrait \cite{wei2024aniportrait}, Echomimic \cite{chen2024echomimic}, and FollowYourEmoji \cite{ma2024follow}.
\subsubsection{Qualitative Evaluation of Self-Reenactment}
The qualitative results of the self-reenactment experiment are presented in Fig. \ref{fig_self}. As shown, methods such as OSFV and TPSMM produce lower image quality, particularly when there is a large pose discrepancy between the driving and source images (e.g., the third and fourth columns of the first, fourth, and fifth rows). These methods suffer from significant background detail loss, facial distortions, and inconsistencies in identity and appearance. Moreover, they exhibit poor perception of depth cues when the subject interacts with objects—for instance, the cup filled with yellow liquid in the second row and the microphone in the third row are both occluded by the face. In comparison, LivePortrait benefits from a high-quality, large-scale dataset of talking heads, enabling improved image generation. It better preserves background details and facial identity, producing high-resolution results. Nevertheless, LivePortrait still faces challenges with occlusions involving objects such as microphones or cups, indicating limitations in handling complex spatial interactions.

Additionally, FADM yields relatively low-quality reenactment results and exhibits limited awareness of occlusions, as shown in the second and third rows of the sixth column. AniPortrait generates visually appealing outputs but struggles to preserve facial identity, particularly in the first and fourth rows of the seventh column. Echomimic shows weaknesses in object perception, leading to artifacts such as missing hair in the first-row image of the eighth column, the disappearance of a cup in the second row, and the occlusion of a microphone by the face in the third row. FollowYourEmoji, meanwhile, reveals mismatches between the reenacted facial poses and those in the driving video, as seen in the first and second rows of the ninth column. Overall, in the qualitative comparison for self-reenactment, our method demonstrates superior performance across several critical dimensions, including image quality, pose accuracy, expression consistency, and robust handling of occlusions involving surrounding objects. These results highlight our method’s consistent advantage over existing state-of-the-art approaches.
\subsubsection{Qualitative Evaluation of Cross-identity Reenactment}
Fig. \ref{fig_cross} presents the qualitative results for the cross-identity reenactment experiment, where the source and driving images belong to different individuals. The objective is to match the facial pose in the driving image while maintaining the identity and appearance of the source. We compare our method with recent state-of-the-art face reenactment approaches, demonstrating its effectiveness in preserving identity and achieving accurate pose transfer.

As observed in the figure, OSFV and TPSMM generally produce lower-quality reenacted images and struggle with generating non-facial elements accurately. For instance, in the reenactment results of the first, second, and fourth rows, these methods introduce severe distortions to headwear. Additionally, in the third row, where the source image contains an occlusion (a hand covering part of the face), they fail to reconstruct the occluded regions properly, leading to severe facial distortions, such as a missing realistic nose and mouth. Similarly, in the fifth row, where the source face is partially obscured by a blue pillar, the reenacted face incorrectly appears in front of the pillar, indicating that these methods struggle to handle occlusions and depth relationships between objects and faces effectively.

In contrast, LivePortrait, FADM, AniPortrait, Echomimic, and FollowYourEmoji generally achieve higher overall image quality than OSFV and TPSMM. However, these methods still face challenges in handling headwear distortions. As observed in the first, second, and fourth rows, although they can synthesize hats, the resulting structures often exhibit visible deformation. When dealing with occlusions, these methods show enhanced reconstruction capabilities for missing facial regions—for instance, in the third row, the occluded nose and mouth are plausibly recovered. Nevertheless, this often comes at the cost of identity inconsistency, where the reenacted face diverges noticeably from the source identity. For the blue pillar occlusion in the fifth row, these methods successfully render the face behind the pillar, preserving correct spatial relationships. However, background artifacts remain an issue—for example, in the seventh column of the fifth row, a black object appears unnaturally where the face should be partially visible, reflecting limitations in background-foreground reasoning.

Overall, our method surpasses state-of-the-art approaches in both image quality and identity preservation, particularly in handling challenges such as headwear deformation, facial occlusions (e.g., hands), and maintaining correct depth ordering in occluded scenes.
\section{Ablation Study}
\subsection{Effectiveness of Warping Feature Mapper}
We validate the effectiveness of the proposed Warping Feature Mapper (WFM) through an ablation study, as shown in Table \ref{tab:abl0}. “w/” denotes the use of WFM, while “w/o” represents the baseline without it. Our method consistently outperforms the baseline across all five evaluation metrics—L1, PSNR, SSIM, LPIPS, and ID. Specifically, L1 and LPIPS are reduced by 81.9\% and 82.2\%, respectively, while PSNR, SSIM, and ID are improved by 88.6\%, 57.9\%, and 60.0\%, demonstrating the significant contribution of WFM to the overall performance.

Without WFM, the SVD model loses the motion constraints provided by WFM, effectively degenerating into a random image-to-video generation model. As a result, the generated outputs deviate significantly from the ground truth, which is clearly reflected in the performance drop shown in Table \ref{tab:abl0}.

\begin{figure}
	\centering
	\includegraphics[width=0.45\textwidth]{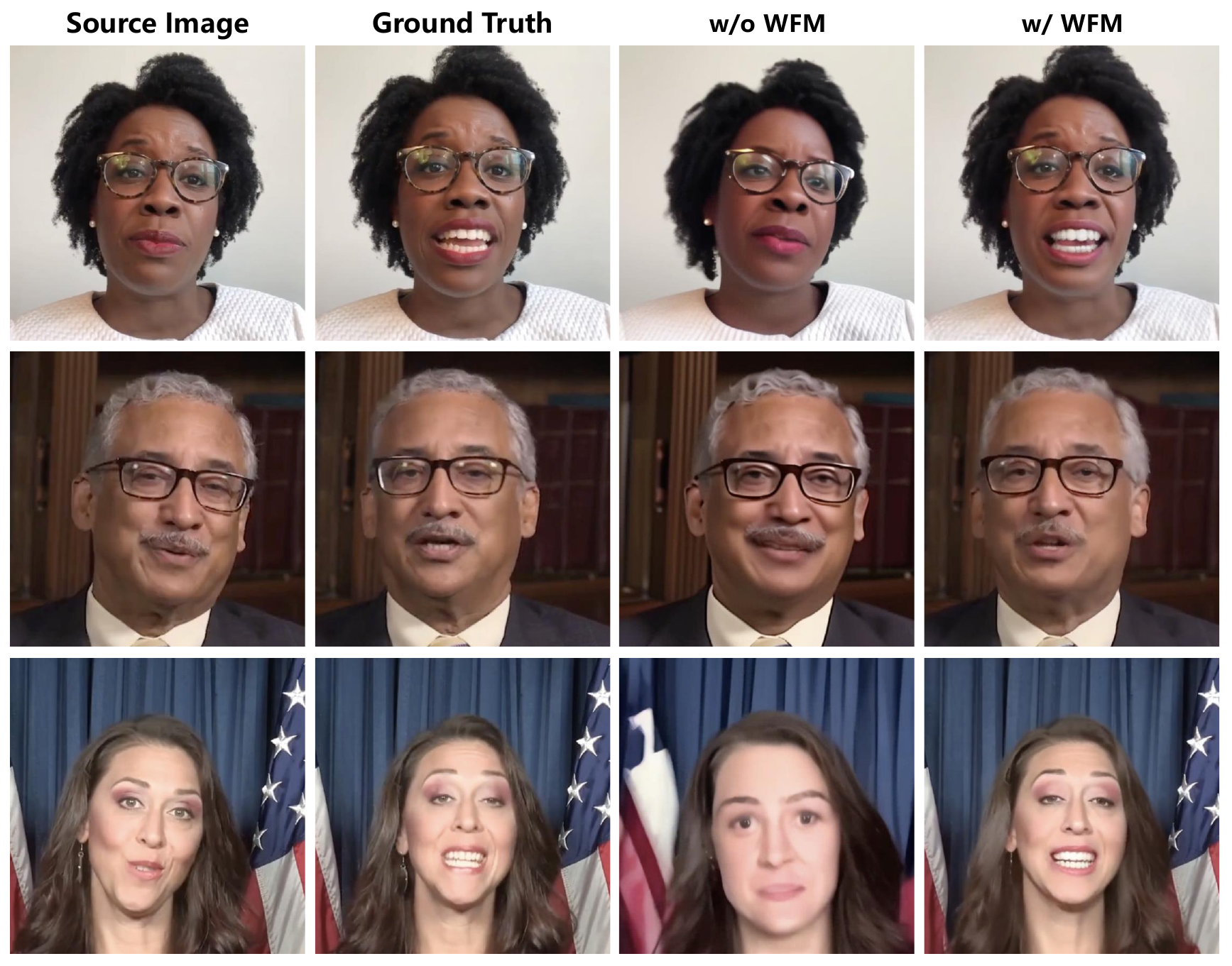}
	\caption{Qualitative results of the ablation study with and without the Warping Feature Mapper (WFM). The first column presents the source image, while the second column shows the ground-truth driving image. The third column illustrates the reenactment results without WFM, and the fourth column shows the results with WFM. Incorporating WFM leads to more realistic and visually faithful reenactments, demonstrating its effectiveness in improving generation quality.}
	\label{fig_abl0}
\end{figure}

We further visualize the reenactment results with and without WFM in Fig. \ref{fig_abl0}. Without WFM, the reenacted faces exhibit misalignment in motion compared to the ground truth and suffer from severe background degradation. For example, in the third column of the first row, the mouth remains closed and the head scale is incorrect; in the second row, the head pose is noticeably wrong; and in the third row, both the facial pose and scale are inaccurate, accompanied by visible background changes. In contrast, with WFM, the reenactment results show more accurate facial pose and scale, as well as better background preservation. This improvement is mainly attributed to WFM’s ability not only to provide motion information to the I2V model but also to spatially constrain the generation process, ensuring that the synthesized faces match the motion patterns of the driving video.
\begin{table}[t!]
    \centering
    \caption{Quantitative results of the ablation study on the Warping Feature Mapper (WFM). “w/” indicates the inclusion of WFM, while “w/o” denotes its removal. The best performance is highlighted in bold, and the second-best is underlined for clarity.}
    \label{tab:abl0}
    \scalebox{0.9}{ 
        \begin{tabular}{cccccc}
            \hline
            Methods & L1$\downarrow$ & PSNR$\uparrow$ & SSIM$\uparrow$ & LPIPS$\downarrow$ & ID$\uparrow$ \\
            \hline
            w/o WFM & \underline{0.1247} & \underline{14.694} & \underline{0.5512} & \underline{0.4270} & \underline{0.5358}\\ 
            \hline
            w/ WFM & \textbf{0.0226} & \textbf{27.708} & \textbf{0.8702} & \textbf{0.0760} & \textbf{0.8570} \\
            \hline
        \end{tabular}
    }
    % \vspace{-4mm}
\end{table}

\subsection{Effectiveness of Rectified Guidance}
\begin{table}[t!]
    \centering
    \caption{Quantitative comparison of the ablation study with rectified guidance used during training. “w/” denotes the use of rectified guidance, while “w/o” indicates its absence. The best scores are highlighted in bold, and the second-best are underlined.}
    \label{tab:abl1}
    \scalebox{0.9}{ 
        \begin{tabular}{ccccc}
            \hline
            Methods & L1$\downarrow$ & PSNR$\uparrow$ & SSIM$\uparrow$ & LPIPS$\downarrow$ \\
            \hline
            w/o rectified guidance & \underline{0.0343} & \underline{25.527} & \underline{0.8295} & \underline{0.0967} \\ 
            \hline
            w/ rectified guidance & \textbf{0.0226} & \textbf{27.708} & \textbf{0.8702} & \textbf{0.0760} \\
            \hline
        \end{tabular}
    }
    % \vspace{-4mm}
\end{table}
\begin{figure}
	\centering
	\includegraphics[width=0.45\textwidth]{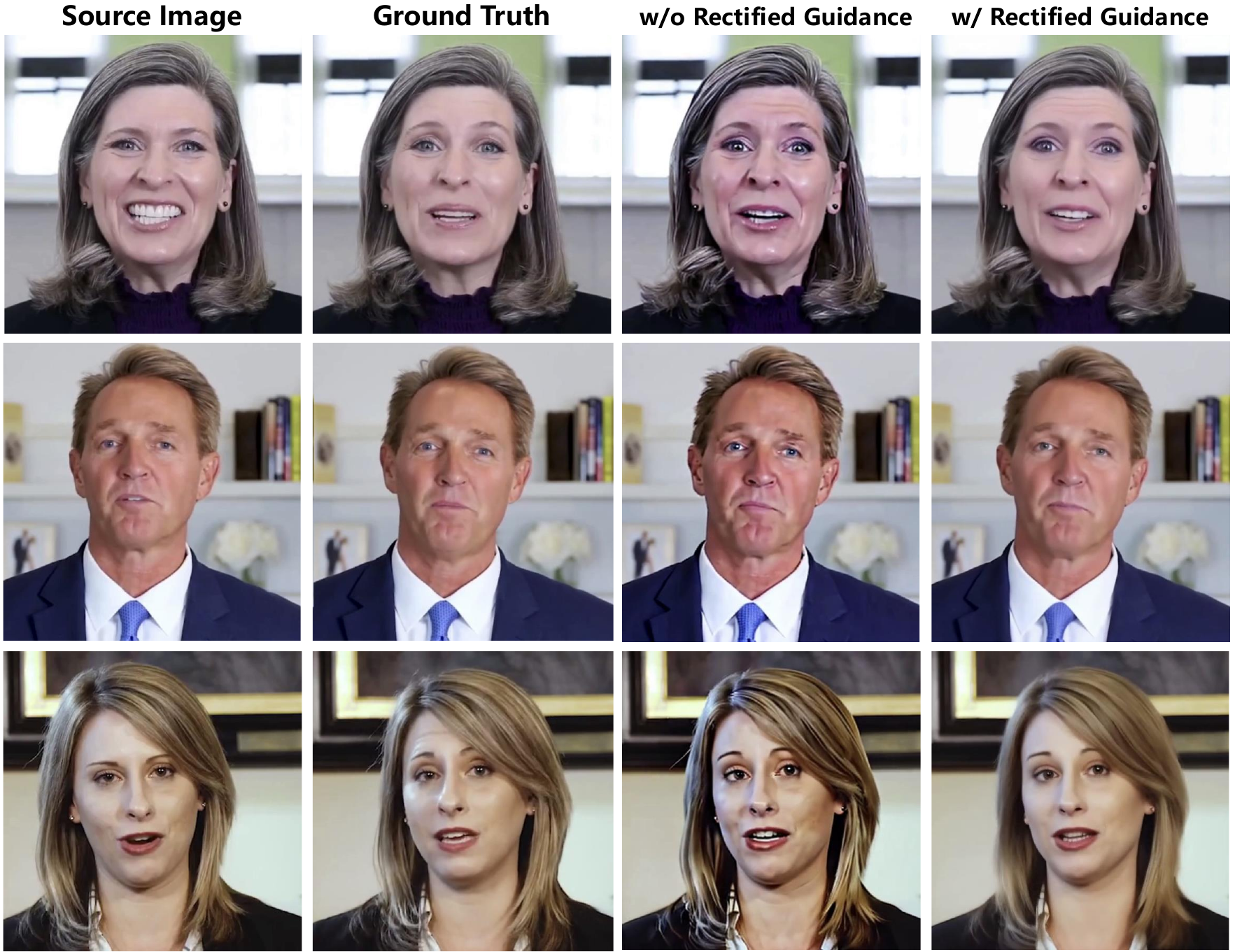}
	\caption{Qualitative comparison of the ablation study with and without rectified guidance during training. The first column shows the source image, and the second column presents the ground truth (driving image). The third column displays the reenactment results without rectified guidance, while the fourth column shows the results with rectified guidance. Our method produces more realistic and visually faithful reenactments when rectified guidance is used.}
	\label{fig_abl1}
\end{figure}
We validate the effectiveness of the proposed rectified guidance, as shown in Table \ref{tab:abl1}. During training, we incorporate rectified guidance into the loss function, denoted as “w/” in the table, while “w/o” indicates training without rectified guidance. Across all four metrics—L1, PSNR, SSIM, and LPIPS—our method achieves consistent improvements. Notably, L1 and LPIPS are reduced by 34.1\% and 21.4\%, respectively, compared to the model trained without rectified guidance.

We further visualize the reenactment results with and without rectified guidance in Fig. \ref{fig_abl1}. When rectified guidance is not used, the reenacted faces exhibit noticeable color deviations. For example, in the third column of the first row, the lips appear overly dark; in the second row, the skin tone deviates significantly; and in the third row, the woman’s hair appears darker than in the source image. In contrast, with rectified guidance, the reenacted results exhibit more natural and faithful appearance, including improved consistency in skin tone, hair color, and lip color, making the overall result more realistic.
\subsection{Effectiveness of Classifier-free Guidance Strength}
\begin{table}[t!]
    \centering
    \caption{Quantitative comparison of the ablation study on classifier-free guidance strength. Different values of w represent varying guidance strengths, with larger w indicating stronger classifier-free guidance. The best scores are highlighted in bold, and the second-best scores are underlined.}
    \label{tab:abl2}
    \scalebox{0.86}{ 
        \begin{tabular}{ccccccc}
            \hline
            Metrics & $w=1$ & $w=2$ & $w=3$ & $w=4$ & $w=5$ & $w=6$\\
            \hline
            L1$\downarrow$ & \underline{0.0235} & \textbf{0.0226} & 0.0289 & 0.0306 & 0.0354 & 0.0471\\ 
            PSNR$\uparrow$ & \underline{27.447} & \textbf{27.708} & 26.167 & 25.854 & 24.883 & 22.786 \\ 
            SSIM$\uparrow$ & \underline{0.8702} & \textbf{0.8730} & 0.8325 & 0.8222 & 0.7853 & 0.7084\\ 
            LPIPS$\downarrow$ & \underline{0.1089} & \textbf{0.0760} & 0.1697 & 0.2271 & 0.3395 & 0.4977 \\ 
            \hline
        \end{tabular}
    }
    % \vspace{-4mm}
\end{table}
\begin{figure}
	\centering
	\includegraphics[width=0.45\textwidth]{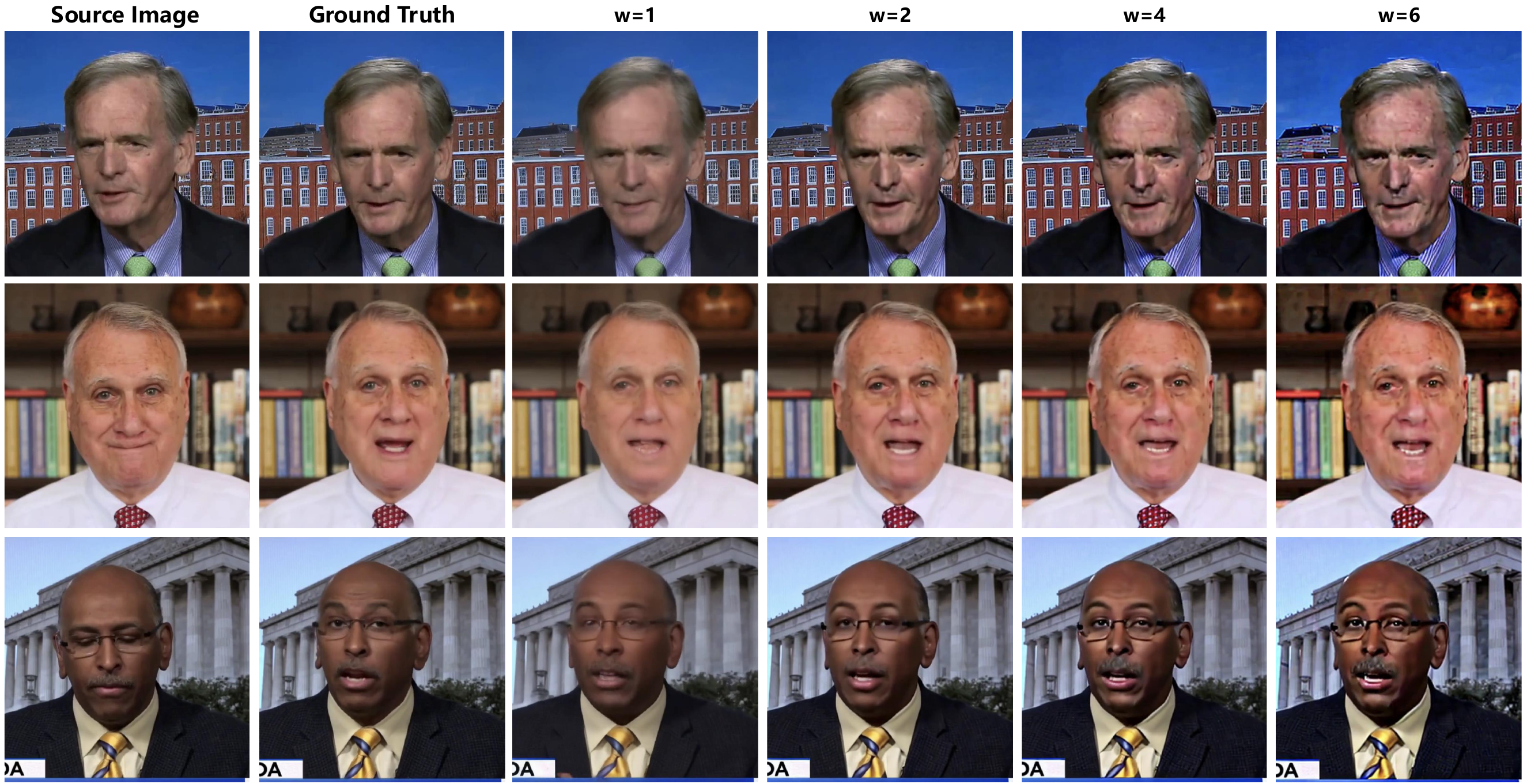}
	\caption{Qualitative comparison of the ablation study on classifier-free guidance strength. The first column shows the source image, and the second column presents the ground truth (driving image). Columns three to six display the reenacted face results under different guidance strengths. As the guidance strength increases (i.e., larger w), facial textures become more pronounced. However, overly strong guidance leads to excessively enhanced high-frequency details, resulting in blocky artifacts on the face.}
	\label{fig_abl2}
\end{figure}
Our method leverages the prior knowledge of a pre-trained SVD model to recover the warped regions of the source image, thereby achieving face reenactment. From the perspective of conditional image generation, the warped image can also be viewed as a condition guiding the SVD model. Therefore, the classifier-free guidance strength (denoted as $w$) directly influences the final reenactment performance.

To validate this, we conduct a comparative study across different w values. As shown in Table \ref{tab:abl2}, when the guidance strength exceeds 1, the performance metrics—L1, PSNR, SSIM, and LPIPS—consistently degrade as $w$ increases. The best results are achieved at $w = 2$. When $w = 1$, the metrics are slightly worse than those at $w = 2$, but still relatively close. We also present qualitative comparisons of facial reenactment under different $w$ values in Fig. \ref{fig_abl2}. When $w = 1$, significant detail loss is observed in the reenacted faces, leading to a hazy or blurred appearance. In contrast, when $w$ exceeds 2, facial details become overly pronounced. For example, in the first and second rows, the age spots on the faces become increasingly blocky and prominent, deviating noticeably from the ground truth, which results in a decline in performance metrics. At $w = 2$, the reenacted faces are visually closest to the ground truth. Overall, classifier-free guidance strength significantly affects the detail level of reenacted faces: higher values enhance detail but may introduce artifacts and distortions if the strength surpasses a certain threshold.
\section{Conclusion}
In this work, we presented FRVD, a novel framework for high-fidelity face reenactment under large pose variations. By leveraging implicit facial keypoints to model fine-grained motion and employing a warping module for motion alignment, our method effectively transfers facial dynamics from the driving video to the source image. To address the degradation introduced by warping, we proposed a Warping Feature Mapper (WFM) that maps the warped source image into the motion-aware latent space of a pretrained image-to-video model. This design enables perceptually accurate reconstruction of facial details and ensures temporal consistency across frames. Extensive experiments demonstrate that FRVD outperforms state-of-the-art methods in terms of pose accuracy, identity preservation, and visual quality, especially in scenarios with extreme head movements. Our approach highlights the potential of combining implicit motion representations with pretrained video priors for robust and expressive face reenactment.

While FRVD demonstrates strong performance in high-fidelity face reenactment under large pose variations, we acknowledge certain limitations. Specifically, the current inference speed (approximately 4 minutes per 100 frames) may hinder real-time applications. In addition, although FRVD effectively handles significant pose changes, further evaluation on extreme non-frontal views (e.g., back-facing poses) is warranted. In future work, we plan to address these issues by exploring model distillation techniques to improve inference efficiency and by extending the evaluation to a broader spectrum of head orientations.
% \section*{Acknowledgments}
% This should be a simple paragraph before the References to thank those individuals and institutions who have supported your work on this article.

%% The Appendices part is started with the command \appendix;
%% appendix sections are then done as normal sections
% \appendix
% \section{Example Appendix Section}
% \label{app1}

% Appendix text.

% %% For citations use: 
% %%       \cite{<label>} ==> [1]

% %%
% Example citation, See \cite{lamport94}.

%% If you have bib database file and want bibtex to generate the
%% bibitems, please use
%%
%%  \bibliographystyle{elsarticle-num} 
%%  \bibliography{<your bibdatabase>}

%% else use the following coding to input the bibitems directly in the
%% TeX file.

%% Refer following link for more details about bibliography and citations.
%% https://en.wikibooks.org/wiki/LaTeX/Bibliography_Management
\bibliographystyle{plain}
\bibliography{reference}

\begin{thebibliography}{10}

\bibitem{blattmann2023stable}
Andreas Blattmann, Tim Dockhorn, Sumith Kulal, Daniel Mendelevitch, Maciej
  Kilian, Dominik Lorenz, Yam Levi, Zion English, Vikram Voleti, Adam Letts,
  et~al.
\newblock Stable video diffusion: Scaling latent video diffusion models to
  large datasets.
\newblock {\em arXiv preprint arXiv:2311.15127}, 2023.

\bibitem{bounareli2024diffusionact}
Stella Bounareli, Christos Tzelepis, Vasileios Argyriou, Ioannis Patras, and
  Georgios Tzimiropoulos.
\newblock Diffusionact: Controllable diffusion autoencoder for one-shot face
  reenactment.
\newblock {\em arXiv preprint arXiv:2403.17217}, 2024.

\bibitem{bulat2017far}
Adrian Bulat and Georgios Tzimiropoulos.
\newblock How far are we from solving the 2d \& 3d face alignment problem? (and
  a dataset of 230,000 3d facial landmarks).
\newblock In {\em International Conference on Computer Vision}, 2017.

\bibitem{chen2024echomimic}
Zhiyuan Chen, Jiajiong Cao, Zhiquan Chen, Yuming Li, and Chenguang Ma.
\newblock Echomimic: Lifelike audio-driven portrait animations through editable
  landmark conditions.
\newblock {\em arXiv preprint arXiv:2407.08136}, 2024.

\bibitem{deng2019arcface}
Jiankang Deng, Jia Guo, Niannan Xue, and Stefanos Zafeiriou.
\newblock Arcface: Additive angular margin loss for deep face recognition.
\newblock In {\em Proceedings of the IEEE/CVF conference on computer vision and
  pattern recognition}, pages 4690--4699, 2019.

\bibitem{esser2021taming}
Patrick Esser, Robin Rombach, and Bjorn Ommer.
\newblock Taming transformers for high-resolution image synthesis.
\newblock In {\em Proceedings of the IEEE/CVF conference on computer vision and
  pattern recognition}, pages 12873--12883, 2021.

\bibitem{goodfellow2014generative}
Ian Goodfellow, Jean Pouget-Abadie, Mehdi Mirza, Bing Xu, David Warde-Farley,
  Sherjil Ozair, Aaron Courville, and Yoshua Bengio.
\newblock Generative adversarial nets.
\newblock {\em Advances in neural information processing systems}, 27, 2014.

\bibitem{guo2024liveportrait}
Jianzhu Guo, Dingyun Zhang, Xiaoqiang Liu, Zhizhou Zhong, Yuan Zhang, Pengfei
  Wan, and Di~Zhang.
\newblock Liveportrait: Efficient portrait animation with stitching and
  retargeting control.
\newblock {\em arXiv preprint arXiv:2407.03168}, 2024.

\bibitem{guo2025high}
Mingtao Guo, Guanyu Xing, and Yanli Liu.
\newblock High-fidelity relightable monocular portrait animation with
  lighting-controllable video diffusion model.
\newblock In {\em Proceedings of the Computer Vision and Pattern Recognition
  Conference}, pages 228--238, 2025.

\bibitem{guoanimatediff}
Yuwei Guo, Ceyuan Yang, Anyi Rao, Zhengyang Liang, Yaohui Wang, Yu~Qiao,
  Maneesh Agrawala, Dahua Lin, and Bo~Dai.
\newblock Animatediff: Animate your personalized text-to-image diffusion models
  without specific tuning.
\newblock In {\em The Twelfth International Conference on Learning
  Representations}.

\bibitem{heusel2017gans}
Martin Heusel, Hubert Ramsauer, Thomas Unterthiner, Bernhard Nessler, and Sepp
  Hochreiter.
\newblock Gans trained by a two time-scale update rule converge to a local nash
  equilibrium.
\newblock {\em Advances in neural information processing systems}, 30, 2017.

\bibitem{hong2022depth}
Fa-Ting Hong, Longhao Zhang, Li~Shen, and Dan Xu.
\newblock Depth-aware generative adversarial network for talking head video
  generation.
\newblock In {\em Proceedings of the IEEE/CVF conference on computer vision and
  pattern recognition}, pages 3397--3406, 2022.

\bibitem{hu2024animate}
Li~Hu.
\newblock Animate anyone: Consistent and controllable image-to-video synthesis
  for character animation.
\newblock In {\em Proceedings of the IEEE/CVF Conference on Computer Vision and
  Pattern Recognition}, pages 8153--8163, 2024.

\bibitem{karras2021alias}
Tero Karras, Miika Aittala, Samuli Laine, Erik H{\"a}rk{\"o}nen, Janne
  Hellsten, Jaakko Lehtinen, and Timo Aila.
\newblock Alias-free generative adversarial networks.
\newblock {\em Advances in Neural Information Processing Systems}, 34:852--863,
  2021.

\bibitem{karras2020analyzing}
Tero Karras, Samuli Laine, Miika Aittala, Janne Hellsten, Jaakko Lehtinen, and
  Timo Aila.
\newblock Analyzing and improving the image quality of stylegan.
\newblock In {\em Proceedings of the IEEE/CVF conference on computer vision and
  pattern recognition}, pages 8110--8119, 2020.

\bibitem{kingma2014adam}
Diederik~P Kingma and Jimmy Ba.
\newblock Adam: A method for stochastic optimization.
\newblock {\em arXiv preprint arXiv:1412.6980}, 2014.

\bibitem{kligvasser2024anchored}
Idan Kligvasser, Regev Cohen, George Leifman, Ehud Rivlin, and Michael Elad.
\newblock Anchored diffusion for video face reenactment.
\newblock {\em arXiv preprint arXiv:2407.15153}, 2024.

\bibitem{krizhevsky2012imagenet}
Alex Krizhevsky, Ilya Sutskever, and Geoffrey~E Hinton.
\newblock Imagenet classification with deep convolutional neural networks.
\newblock {\em Advances in neural information processing systems}, 25, 2012.

\bibitem{lugaresi2019mediapipe}
Camillo Lugaresi, Jiuqiang Tang, Hadon Nash, Chris McClanahan, Esha Uboweja,
  Michael Hays, Fan Zhang, Chuo-Ling Chang, Ming~Guang Yong, Juhyun Lee, et~al.
\newblock Mediapipe: A framework for building perception pipelines.
\newblock {\em arXiv preprint arXiv:1906.08172}, 2019.

\bibitem{ma2024follow}
Yue Ma, Hongyu Liu, Hongfa Wang, Heng Pan, Yingqing He, Junkun Yuan, Ailing
  Zeng, Chengfei Cai, Heung-Yeung Shum, Wei Liu, et~al.
\newblock Follow-your-emoji: Fine-controllable and expressive freestyle
  portrait animation.
\newblock In {\em SIGGRAPH Asia 2024 Conference Papers}, pages 1--12, 2024.

\bibitem{rombach2022high}
Robin Rombach, Andreas Blattmann, Dominik Lorenz, Patrick Esser, and Bj{\"o}rn
  Ommer.
\newblock High-resolution image synthesis with latent diffusion models.
\newblock In {\em Proceedings of the IEEE/CVF conference on computer vision and
  pattern recognition}, pages 10684--10695, 2022.

\bibitem{siarohin2019first}
Aliaksandr Siarohin, St{\'e}phane Lathuili{\`e}re, Sergey Tulyakov, Elisa
  Ricci, and Nicu Sebe.
\newblock First order motion model for image animation.
\newblock {\em Advances in neural information processing systems}, 32, 2019.

\bibitem{song2020denoising}
Jiaming Song, Chenlin Meng, and Stefano Ermon.
\newblock Denoising diffusion implicit models.
\newblock In {\em International Conference on Learning Representations}, 2020.

\bibitem{unterthiner2018towards}
Thomas Unterthiner, Sjoerd van Steenkiste, Karol Kurach, Raphael Marinier,
  Marcin Michalski, and Sylvain Gelly.
\newblock Towards accurate generative models of video: A new metric \&
  challenges.
\newblock {\em arXiv preprint arXiv:1812.01717}, 2018.

\bibitem{wan2025}
Ang Wang, Baole Ai, Bin Wen, Chaojie Mao, Chen-Wei Xie, Di~Chen, Feiwu Yu,
  Haiming Zhao, Jianxiao Yang, Jianyuan Zeng, Jiayu Wang, Jingfeng Zhang,
  Jingren Zhou, Jinkai Wang, Jixuan Chen, Kai Zhu, Kang Zhao, Keyu Yan,
  Lianghua Huang, Mengyang Feng, Ningyi Zhang, Pandeng Li, Pingyu Wu, Ruihang
  Chu, Ruili Feng, Shiwei Zhang, Siyang Sun, Tao Fang, Tianxing Wang, Tianyi
  Gui, Tingyu Weng, Tong Shen, Wei Lin, Wei Wang, Wei Wang, Wenmeng Zhou, Wente
  Wang, Wenting Shen, Wenyuan Yu, Xianzhong Shi, Xiaoming Huang, Xin Xu, Yan
  Kou, Yangyu Lv, Yifei Li, Yijing Liu, Yiming Wang, Yingya Zhang, Yitong
  Huang, Yong Li, You Wu, Yu~Liu, Yulin Pan, Yun Zheng, Yuntao Hong, Yupeng
  Shi, Yutong Feng, Zeyinzi Jiang, Zhen Han, Zhi-Fan Wu, and Ziyu Liu.
\newblock Wan: Open and advanced large-scale video generative models.
\newblock {\em arXiv preprint arXiv:2503.20314}, 2025.

\bibitem{wang2021facevid2vid}
Ting-Chun Wang, Arun Mallya, and Ming-Yu Liu.
\newblock One-shot free-view neural talking-head synthesis for video
  conferencing.
\newblock In {\em Proceedings of the IEEE Conference on Computer Vision and
  Pattern Recognition}, 2021.

\bibitem{wang2004image}
Zhou Wang, Alan~C Bovik, Hamid~R Sheikh, and Eero~P Simoncelli.
\newblock Image quality assessment: from error visibility to structural
  similarity.
\newblock {\em IEEE transactions on image processing}, 13(4):600--612, 2004.

\bibitem{wei2024aniportrait}
Huawei Wei, Zejun Yang, and Zhisheng Wang.
\newblock Aniportrait: Audio-driven synthesis of photorealistic portrait
  animations, 2024.

\bibitem{wu2022fasterquality}
Haoning Wu, Chaofeng Chen, Liang Liao, Jingwen Hou, Wenxiu Sun, Qiong Yan,
  Jinwei Gu, and Weisi Lin.
\newblock Neighbourhood representative sampling for efficient end-to-end video
  quality assessment, 2022.

\bibitem{xie2022vfhq}
Liangbin Xie, Xintao Wang, Honglun Zhang, Chao Dong, and Ying Shan.
\newblock Vfhq: A high-quality dataset and benchmark for video face
  super-resolution.
\newblock In {\em Proceedings of the IEEE/CVF Conference on Computer Vision and
  Pattern Recognition}, pages 657--666, 2022.

\bibitem{xie2024x}
You Xie, Hongyi Xu, Guoxian Song, Chao Wang, Yichun Shi, and Linjie Luo.
\newblock X-portrait: Expressive portrait animation with hierarchical motion
  attention.
\newblock In {\em ACM SIGGRAPH 2024 Conference Papers}, pages 1--11, 2024.

\bibitem{yang2024cogvideox}
Zhuoyi Yang, Jiayan Teng, Wendi Zheng, Ming Ding, Shiyu Huang, Jiazheng Xu,
  Yuanming Yang, Wenyi Hong, Xiaohan Zhang, Guanyu Feng, et~al.
\newblock Cogvideox: Text-to-video diffusion models with an expert transformer.
\newblock {\em arXiv preprint arXiv:2408.06072}, 2024.

\bibitem{zeng2023face}
Bohan Zeng, Xuhui Liu, Sicheng Gao, Boyu Liu, Hong Li, Jianzhuang Liu, and
  Baochang Zhang.
\newblock Face animation with an attribute-guided diffusion model.
\newblock In {\em Proceedings of the IEEE/CVF Conference on Computer Vision and
  Pattern Recognition}, pages 628--637, 2023.

\bibitem{zhang2018unreasonable}
Richard Zhang, Phillip Isola, Alexei~A Efros, Eli Shechtman, and Oliver Wang.
\newblock The unreasonable effectiveness of deep features as a perceptual
  metric.
\newblock In {\em Proceedings of the IEEE conference on computer vision and
  pattern recognition}, pages 586--595, 2018.

\bibitem{zhang2024mimicmotion}
Yuang Zhang, Jiaxi Gu, Li-Wen Wang, Han Wang, Junqi Cheng, Yuefeng Zhu, and
  Fangyuan Zou.
\newblock Mimicmotion: High-quality human motion video generation with
  confidence-aware pose guidance.
\newblock {\em arXiv preprint arXiv:2406.19680}, 2024.

\bibitem{zhang2021flow}
Zhimeng Zhang, Lincheng Li, Yu~Ding, and Changjie Fan.
\newblock Flow-guided one-shot talking face generation with a high-resolution
  audio-visual dataset.
\newblock In {\em Proceedings of the IEEE/CVF conference on computer vision and
  pattern recognition}, pages 3661--3670, 2021.

\bibitem{zhao2022thin}
Jian Zhao and Hui Zhang.
\newblock Thin-plate spline motion model for image animation.
\newblock In {\em Proceedings of the IEEE/CVF Conference on Computer Vision and
  Pattern Recognition}, pages 3657--3666, 2022.

\bibitem{zhu2022celebv}
Hao Zhu, Wayne Wu, Wentao Zhu, Liming Jiang, Siwei Tang, Li~Zhang, Ziwei Liu,
  and Chen~Change Loy.
\newblock Celebv-hq: A large-scale video facial attributes dataset.
\newblock In {\em European conference on computer vision}, pages 650--667.
  Springer, 2022.

\end{thebibliography}
% \begin{thebibliography}{00}

% %% For numbered reference style
% %% \bibitem{label}
% %% Text of bibliographic item

% \bibitem{siarohin2019first}
%   Aliaksandr Siarohin, Stéphane Lathuilière, Sergey Tulyakov, Elisa Ricci, and Nicu Sebe,
%   \textit{First order motion model for image animation},
%   Advances in neural information processing systems,
%   volume 32,
%   2019.

% \bibitem{lamport94}
%   Leslie Lamport,
%   \textit{\LaTeX: a document preparation system},
%   Addison Wesley, Massachusetts,
%   2nd edition,
%   1994.

% \end{thebibliography}
\end{document}